\documentclass[10pt,twocolumn,letterpaper]{article}

\usepackage{cvpr}              % To produce the CAMERA-READY version
\usepackage{multirow}
\usepackage{wrapfig}
\usepackage{adjustbox}
\usepackage[dvipsnames]{xcolor}

\definecolor{cvprblue}{rgb}{0.21,0.49,0.74}
\usepackage[pagebackref,breaklinks,colorlinks,citecolor=cvprblue]{hyperref}

\def\paperID{*****} % *** Enter the Paper ID here
\def\confName{CVPR}
\def\confYear{2024}

\title{SAM3-Adapter: Efficient Adaptation of Segment Anything 3 for Camouflage Object Segmentation, Shadow Detection, and Medical Image Segmentation}

\author{Tianrun Chen$^{1,2*+}$\and Runlong Cao$^{3+}$\and Xinda Yu$^{4+}$ \and Lanyun Zhu$^{5}$ \and Chaotao Ding$^{1}$ \\ \and Deyi Ji $^{7}$ \and Cheng Chen$^{6}$ \and Qi Zhu$^{7}$ \and Chunyan Xu$^{3}$ \and Papa Mao$^{1}$ \\\and Ying Zang$^{4*}$\\
\\
\textbf{This work is based on the \href{http://tianrun-chen.github.io/SAM-Adaptor/}{SAM-Adapter}, which was originally released on April 14, 2023.} \\
\textbf{TL; DR}. SAM3, enhanced with our proposed adapter, surpasses its predecessor \\ as a backbone for segmentation and establishes new state-of-the-art (SOTA) results \\ across a range of downstream tasks \\
\\
\small $^+$ Equal Contribution $^*$ \small Corresponding Author \small$\{$tianrun.chen@zju.edu.cn; 02750@zjhu.edu.cn$\}$
\\
\small $^{1}$KOKONI, Moxin (Huzhou) Tech. Co., LTD, Huzhou, Zhejiang, P.R. China.\\ 
\small $^{2}$College of Computer Science and Technology, Zhejiang University, Hangzhou, Zhejiang, P.R. China.\\ 
\small $^{3}$School of Computer Science and Engineering, Nanjing University of Science and Technology, Nanjing, P.R. China.\\ 
\small $^{4}$School of Information Engineering, Huzhou University, Huzhou, P.R. China.\\ 
\small $^{5}$ School of Electrical and Electronic Engineering, Nanyang Technological University, Singapore.\\
\small $^{6}$ College of Computing and Data Science, Nanyang Technological University, Singapore.\\ 
\small $^{7}$ School of Information Science and Technology, University of Science and Technology of China, P.R. China.\\ 
\\
Project Page: \href{http://tianrun-chen.github.io/SAM-Adaptor/}{http://tianrun-chen.github.io/SAM-Adaptor/}
}

\begin{document}
\maketitle
\begin{abstract}
   The rapid rise of large-scale foundation models has reshaped the landscape of image segmentation, with models such as Segment Anything achieving unprecedented versatility across diverse vision tasks. However, previous generations—including SAM and its successor—still struggle with fine-grained, low-level segmentation challenges such as camouflaged object detection, medical image segmentation, cell image segmentation, and shadow detection. To address these limitations, we originally proposed SAM-Adapter in 2023, demonstrating substantial gains on these difficult scenarios. With the emergence of Segment Anything 3 (SAM3)—a more efficient and higher-performing evolution with a redesigned architecture and improved training pipeline—we revisit these long-standing challenges. In this work, we present SAM3-Adapter, the first adapter framework tailored for SAM3 that unlocks its full segmentation capability. SAM3-Adapter not only reduces computational overhead but also consistently surpasses both SAM and SAM2-based solutions, establishing new state-of-the-art results across multiple downstream tasks, including medical imaging, camouflaged (concealed) object segmentation, and shadow detection. Built upon the modular and composable design philosophy of the original SAM-Adapter, SAM3-Adapter provides stronger generalizability, richer task adaptability, and significantly improved segmentation precision. Extensive experiments confirm that integrating SAM3 with our adapter yields superior accuracy, robustness, and efficiency compared to all prior SAM-based adaptations. We hope SAM3-Adapter can serve as a foundation for future research and practical segmentation applications. Code, pre-trained models, and data processing pipelines are available at:
\href{http://tianrun-chen.github.io/SAM-Adaptor/}{http://tianrun-chen.github.io/SAM-Adaptor/}.
\end{abstract}    
\section{Introduction}
\label{sec:intro}
The AI research landscape has been revolutionized by foundation models trained on vast datasets \cite{bommasani2021opportunities, zhu2024llafs, zhu2024ibd, chen2024reasoning3d}. Among these, the Segment Anything (SAM) series \cite{SAM} has become a cornerstone for image segmentation. Our prior work, SAM-Adapter \cite{chen2023samfailssegmentanything, chen2023sam} and SAM2-Adapter \cite{chen2024sam2} were pioneering efforts to bridge the gap between the first SAM's general capabilities and the nuanced demands of downstream tasks, a contribution widely adopted by the community.

Now, the release of Segment Anything 3 (SAM3) marks a new era. With its significantly scaled-up architecture and a more extensive training corpus, SAM3 offers a vastly superior foundation and unprecedented potential for segmentation tasks. This advancement shifts the fundamental research question from "how do we fix model limitations?" to "how do we fully unlock and channel the immense power of this scaled-up model for specialized applications?"

This paper provides a definitive answer. We introduce SAM3-Adapter, a highly efficient and synergistic adaptation method designed specifically to unleash the full capabilities of SAM3. We demonstrate that by pairing the powerful SAM3 backbone with our lightweight adapter, we can achieve new state-of-the-art (SOTA) performance on a diverse set of challenging downstream tasks, including medical image, camouflage, and shadow segmentation. This work is the first to prove that the scaling of SAM3, when properly harnessed, directly translates into breakthrough performance in these specialized domains.

Our SAM3-Adapter is engineered to be both Generalizable and Composable. It can be seamlessly applied to custom datasets with minimal data, and its components can be flexibly combined to meet diverse task requirements. Critically, it is tailored to SAM3's advanced hierarchical architecture, ensuring that every part of the powerful backbone is effectively utilized. This synergy allows SAM3-Adapter to not only achieve superior accuracy but also maintain remarkable parameter efficiency.
We conduct extensive experiments on multiple benchmarks, including ISTD \cite{wang2018stacked}, COD10K \cite{fan2020camouflaged}, and Kvasir-SEG \cite{jha2020kvasir}. The results are unequivocal: the combination of SAM3 and SAM3-Adapter consistently outperforms all previous methods. Our contributions are:

\begin{itemize}
    \item We are the first to demonstrate and unlock the latent potential of the scaled-up SAM3 model for specialized downstream tasks, showing that its advanced architecture provides a superior foundation for achieving SOTA performance.
    \item We propose SAM3-Adapter, a novel, parameter-efficient adaptation framework specifically designed to synergize with SAM3, effectively channeling its generalist power into specialist excellence.
    \item We establish new state-of-the-art results across multiple challenging segmentation benchmarks, proving that a generalist backbone like SAM3, when enhanced by our adapter, can outperform highly specialized models.
\end{itemize}

We advocate for adopting the SAM3 and SAM3-Adapter combination to push the frontiers of image segmentation in both research and industry. We encourage the research community to adopt SAM3 as the backbone in conjunction with our SAM3-Adapter, to achieve even better segmentation results in various research fields and industrial applications. We are releasing our code, pre-trained model, and data processing protocols in \href{http://tianrun-chen.github.io/SAM-Adaptor/}{http://tianrun-chen.github.io/SAM-Adaptor/}.

\section{Related Work}
    \noindent \textbf{Semantic Segmentation.} In recent years, semantic segmentation has made significant progress, primarily due to the remarkable advancements in deep-learning-based methods such as fully convolutional networks (FCN) \cite{fcn}, encoder-decoder structures \cite{unet, bisenetv2, segnet, dlpl, chen2024xlstm, urur, sstkd}, dilated convolutions \cite{deeplabv3, deeplabv3+, liu2021label, zang2024resmatch, cdgc}, pyramid structures \cite{zhu2021learning, deeplabv3, pspnet, deeplabv3+, zhu2023continual, fu2022panoptic, cagcn, wang2023fvp}, attention modules \cite{ann, zhu2024addressing, zhu2023learning, pptformer, gpwformer}, and transformers \cite{zheng2021rethinking, xie2021segformer, strudel2021segmenter, cheng2022masked, zhu2024llafs}. Recent advancements have improved SAM's performance, such as \cite{HQ-SAM}, which introduces a High-Quality output token and trains the model on fine-grained masks. Other efforts have focused on enhancing SAM's efficiency for broader real-world and mobile use, exemplified by \cite{EfficientSAM, MobileSAM, FastSAM}. The widespread success of SAM has led to its adoption in various fields, including medical imaging \cite{Ma_2024, deng2023segmentmodelsamdigital, Mazurowski_2023, wu2023medicalsamadapteradapting, ma2024segment}, remote sensing \cite{chen2023rsprompterlearningpromptremote, ren2023segmentanythingspace, changenet}, motion segmentation \cite{xie2024movingobjectsegmentationneed, wang2021learning, ipgn, feng2018challenges}, and camouflaged object detection \cite{tang2023samsegmentanythingsam}. Notably, our previous work SAM-Adapter \cite{chen2023samfailssegmentanything, chen2023sam} tested camouflaged object detection, polyp segmentation, and shadow segmentation, and provide the first adapter-based method to integrate the SAM's exceptional capability to these downstream tasks. \

    \noindent \textbf{Adapters. }The concept of Adapters was first introduced in the NLP community \cite{houlsby2019parameter} as a tool to fine-tune a large pre-trained model for each downstream task with a compact and scalable model. In \cite{stickland2019bert}, multi-task learning was explored with a single BERT model shared among a few task-specific parameters. In the computer vision community, \cite{li2022exploring, homview, mscnn} suggested fine-tuning the ViT \cite{dosovitskiy2020image} for object detection with minimal modifications. Recently, ViT-Adapter \cite{chen2022vision} leveraged Adapters to enable a plain ViT to perform various downstream tasks. \cite{liu2023explicit} introduce an Explicit Visual Prompting (EVP) technique that can incorporate explicit visual cues to the Adapter. However, no prior work has tried to apply Adapters to leverage pretrained image segmentation model SAM trained at large image corpus. Here, we mitigate the research gap. 

\begin{figure*}[t]
\centering
\includegraphics[width=\linewidth]{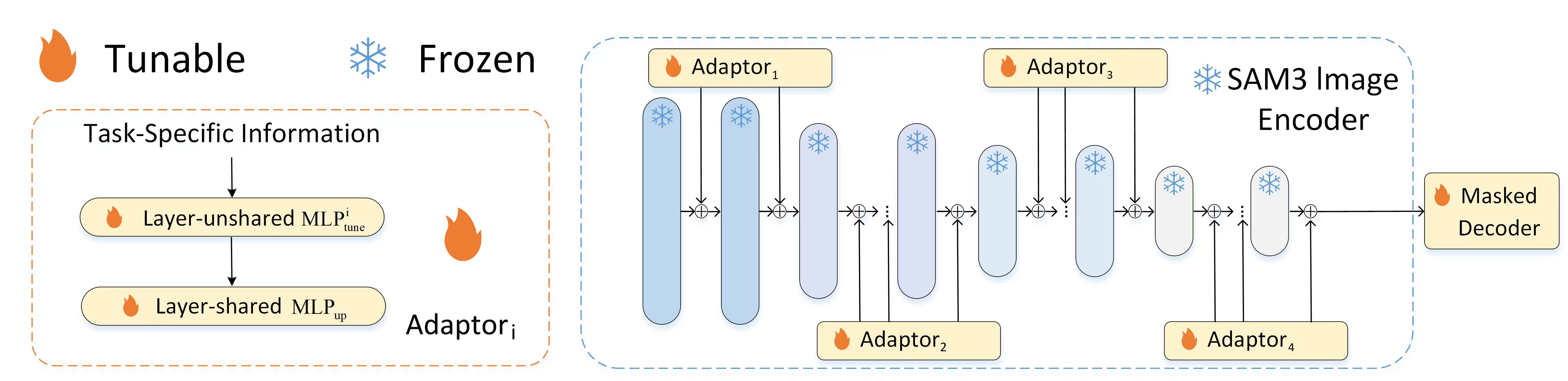}
\caption{\textbf{The architecture of the proposed SAM3-Adapter. }}
\label{framework}
\end{figure*}

\noindent \textbf{Polyp Segmentation.}
    In recent years, there has been notable progress in polyp segmentation \cite{zhou2021fullattentionbasedneuralarchitecture} due to deep-learning approaches. These techniques employ deep neural networks to derive more discriminative features from endoscopic polyp images. Nonetheless, the use of bounding-box detectors often leads to inaccurate polyp boundary localization. To resolve this, \cite{canny1986computational} leveraged fully convolutional networks (FCN) with pre-trained models to identify and segment polyps. \cite{qadir2021toward} introduced a technique utilizing Fully Convolutional Neural Networks (FCNNs) to predict 2D Gaussian shapes. Subsequently, the U-Net \cite{kingma2017adammethodstochasticoptimization} architecture, featuring a contracting path for context capture and a symmetric expanding path for precise localization, achieved favorable segmentation results. However, these strategies focus primarily on entire polyp regions, neglecting boundary constraints. Therefore, Psi-Net \cite{murugesan2019psinetshapeboundaryaware} incorporated both region and boundary constraints for polyp segmentation, yet the interplay between regions and boundaries remained underexplored. \cite{mahmud2021polypsegnet} introduced PolypSegNet, an enhanced encoder-decoder architecture designed for the automated segmentation of polyps in colonoscopy images. To address the issue of non-equivalent images and pixels, \cite{guo2022non} proposed a confidence-aware resampling method for polyp segmentation tasks. Specifically for polyp segmentation, works done by \cite{zhou2023samsegmentpolyps} and \cite{chen2023samfailssegmentanything} present promising results using an unprompted SAM and a domain-adapted SAM respectively. Additionally, Polyp-SAM \cite{li2023polypsamtransfersampolyp} used SAM for the same task. \cite{roy2023sammdzeroshotmedicalimage} evaluated the zero-shot capabilities of SAM on the organ segmentation task.

    \noindent \textbf{Camouflaged Object Detection (COD).} Camouflaged object detection, or concealed object detection is a challenging but useful task that identifies objects that blend in with their surroundings. COD has wide applications in medicine, agriculture, and art. Initially, research of camouflage detection relied on low-level features like texture, brightness, and color \cite{feng2013camouflage,pike2018quantifying,hou2011detection,sengottuvelan2008performance} to distinguish foreground from background. It is worth noting that some of this prior knowledge is critical in identifying the objects, and is used to guide the neural network in this paper.

    Le et al.\cite{le2019anabranch} first proposed an end-to-end network consisting of a classification and a segmentation branch. Recent advances in deep learning-based methods have shown a superior ability to detect complex camouflaged objects \cite{fan2020camouflaged, mei2021camouflaged, lin2023frequency}. In this work, we leverage the advanced neural network backbone (a foundation model -- SAM2) with the input of task-specific prior knowledge to achieve state-of-the-art (SOTA) performance. 

    \noindent \textbf{Shadow Detection.} Shadows can occur when an object's surface is not directly exposed to light. They offer hints on light source direction and scene illumination that can aid scene comprehension \cite{karsch2011rendering,lalonde2012estimating}. They can also negatively impact the performance of computer vision tasks \cite{nadimi2004physical,cucchiara2003detecting}. Early methods use hand-crafted heuristic cues like chromaticity, intensity, and texture \cite{huang2011characterizes,lalonde2012estimating,zhu2010learning}. Deep learning approaches leverage the knowledge learned from data and use delicately designed neural network structures to capture the information (e.g. learned attention modules) \cite{le2018a+,cun2020towards,zhu2018bidirectional}. This work leverages the heuristic priors with large neural network models to achieve the state-of-the-art (SOTA) performance.

\section{Method}

\subsection{Using SAM 3 as the Backbone}
The core of our approach is built upon the formidable vision backbone of the SAM3 model. SAM3 represents a significant architectural evolution, featuring a unified backbone shared between a DETR-based detector and a video tracker, designed to process complex visual and concept-based prompts.

In our work, we leverage this powerful, pre-trained vision encoder from SAM3. We keep its weights frozen during training. This strategy is crucial as it preserves the incredibly rich and generalizable visual representations learned from SAM3's extensive training on the massive SA-Co dataset. By doing so, we build upon a superior foundation without incurring the prohibitive costs of re-training the entire model. For the segmentation head, we utilize the mask decoder architecture from the SAM family, initializing it with pre-trained weights and subsequently fine-tuning it alongside our adapter.

While the SAM3 encoder provides a state-of-the-art foundation, unlocking its full potential for specialized domains requires a mechanism to inject task-specific knowledge. Following the successful paradigm of our previous work \cite{chen2023samfailssegmentanything}, we introduce lightweight adapters to achieve this.

\subsection{SAM3-Adapter for Task-Specific Specialization}
The architecture of our SAM3-Adapter is designed for simplicity and efficiency, as illustrated in Figure \ref{framework}. The SAM3 vision encoder features a hierarchical, multi-stage architecture. To complement this, we introduce a set of adapters, one for each stage of the encoder. The weights of the adapter are shared within each stage to maintain parameter efficiency.

Specifically, each adapter processes task-specific information, $F_i$, to generate a conditioning prompt, $P_i$. This process is defined as:
\begin{equation}
\label{get_context}
P^i = {\rm MLP}{up}\left({\rm GELU}\left({\rm MLP}{tune}^i\left(F_i\right)\right)\right)
\end{equation}

where ${\rm MLP}{tune}^i$ is a tunable linear layer that creates a task-specific prompt from the input information $F_i$. The ${\rm MLP}{up}$ is an up-projection layer, shared across all adapters, that aligns the prompt's dimensions with the transformer features. The resulting prompt $P^i$ is then integrated into the transformer layers of the corresponding stage, effectively guiding the model's focus toward task-relevant features.

\subsection{Flexible Task-Specific Inputs}

A key strength of our framework is the flexibility of the task-specific information, $F_i$. This input can be engineered in various forms depending on the downstream application. For instance, it can be derived from dataset-specific statistics (e.g., texture, frequency information) or based on hand-crafted rules relevant to the task.

Furthermore, $F_i$ can be a composition of multiple guidance signals, allowing for nuanced control:
\begin{align}
F_i & = \sum_{j=1}^{N}w_jF_j
\end{align}

Here, each $F_j$ represents a distinct type of knowledge or feature, and $w_j$ is a learnable weight controlling its influence. This composability enables the model to integrate diverse sources of information, enhancing its adaptability. For a more detailed exploration of this concept, we refer readers to our original SAM-Adapter paper \cite{chen2023samfailssegmentanything}.
\section{Experiments}
\subsection{Tasks and Datasets}
    In our experiments, we selected two challenging low-level structural segmentation tasks and two medical imaging task to evaluate the performance of the SAM2-Adapter: camouflaged object detection and shadow detection, polyp segmentation and cell segmentation.

\begin{figure*}[h]
\centering
\includegraphics[width=0.6\linewidth]{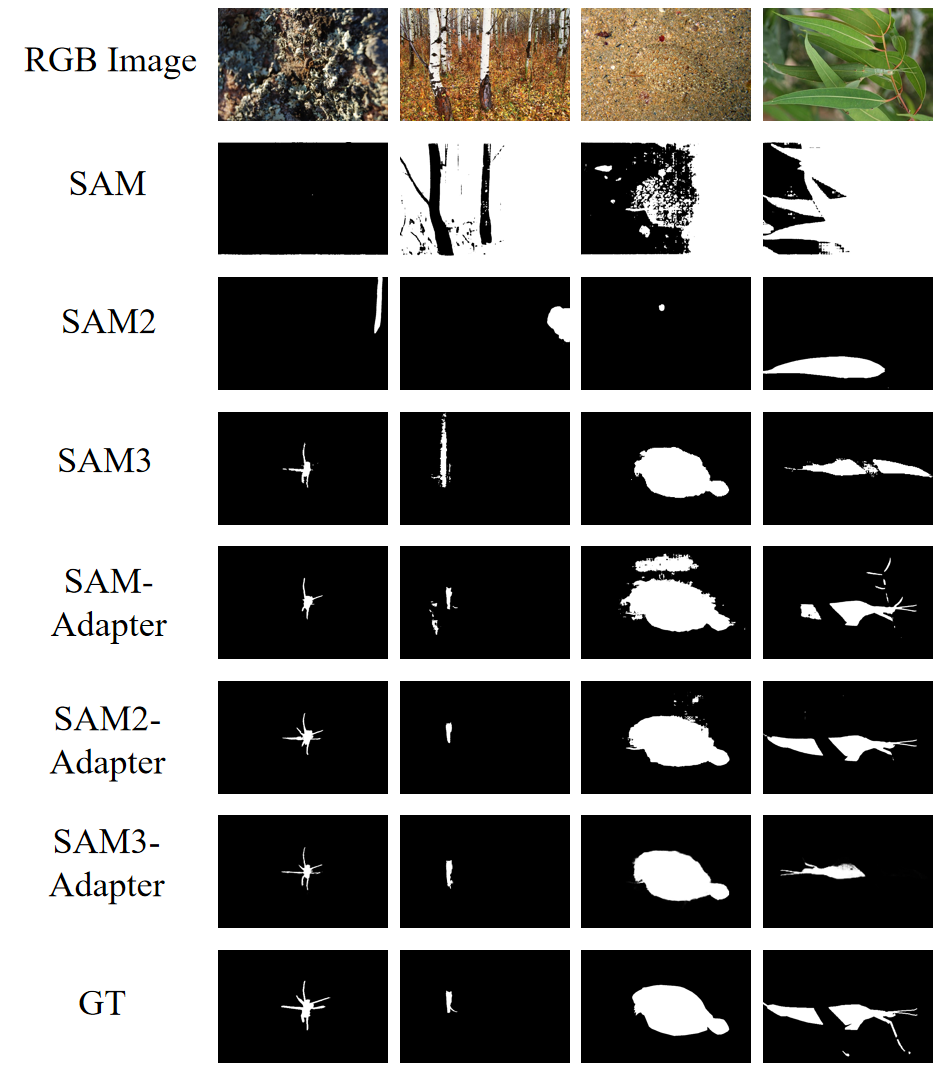}
\caption{\textbf{Segmentation performance visualization on CHAMELEON samples.} The figure highlights the limitations of SAM, SAM2, and SAM3 in handling severe camouflage, where they produce non-meaningful outcomes. Although SAM-Adapter enhances segmentation quality, our SAM3-Adapter delivers the best performance, generating precise masks that closely match the ground truth compared to its predecessors.}
\label{fig2}
\end{figure*}

    For the camouflaged object detection task, we use three widely used benchmarks: COD10K \cite{fan2020camouflaged}, CHAMELEON \cite{skurowski2018animal}, and CAMO \cite{le2019anabranch}. COD10K is currently the most comprehensive resource in this area, offering 3,040 samples for training and 2,026 for testing. The CHAMELEON set provides 76 internet-sourced images and is used solely for evaluation. The CAMO dataset comprises 1,250 images, with 1,000 designated for training and 250 for testing. Following the training strategy proposed in \cite{fan2020camouflaged}, our model was trained using CAMO together with the COD10K training split, while performance was assessed on the test splits of CAMO and COD10K, as well as the full CHAMELEON dataset.

For the shadow detection task, we adopted the ISTD dataset \cite{wang2018stacked}, which includes 1,330 training and 540 testing samples. For polyp segmentation in medical imaging, we used the Kvasir-SEG dataset \cite{jha2020kvasir}, following the train–test division specified in the Medico 2020 Automatic Polyp Segmentation challenge \cite{jha2020medico}. In cell segmentation, we obtain dataset from NeurIPS 2022 Cell Segmentation Challenge \cite{ma2024multimodality}, which focuses on cell segmentation in various microscopy images. We used 1000 images for training and 101
 images for evaluation. The original task was an instance segmentation task. In our experiments, we converted instance segmentation into a semantic segmentation task following the data processing method described in \cite{ma2024u}. 

For the evaluation protocol, we followed the guidelines in \cite{liu2023explicit}. For camouflaged object detection, we adopted widely used metrics, including S-measure ($S_m$), mean E-measure ($E_\phi$), and MAE. The shadow detection task was assessed using the balanced error rate (BER). For polyp segmentation, model performance was quantified using the mean Dice score (mDice) and mean Intersection-over-Union (mIoU).

    For more details, please refer to the original SAM-Adapter paper \cite{chen2023samfailssegmentanything}.

\subsection{Implementation Details}
    In the experiment, we choose two types of visual knowledge, patch embedding $F_{pe}$ and high-frequency components $F_{hfc}$, following the same setting in \cite{liu2023explicit}, which has been demonstrated effective in various of vision tasks. $w^j$ is set to 1. Therefore, the $F_i$ is derived by $F_i=F_{hfc}+F_{pe}$. 

\begin{table*}[t]
\scalebox{0.9}{
\begin{tabular}{c||cccc|cccc|cccc}
\hline
\multirow{2}{*}{Method} & \multicolumn{4}{c|}{CHAMELEON  \cite{skurowski2018animal}}                                    & \multicolumn{4}{c|}{CAMO \cite{le2019anabranch}}                                         & \multicolumn{4}{c}{COD10K \cite{fan2020camouflaged}}                                        \\ \cline{2-13} 
& $ S_\alpha \uparrow$               & $E_\phi \uparrow$              & $F^\omega_\beta \uparrow$              & MAE $\downarrow$           & $S_\alpha \uparrow$              & $E_\phi \uparrow$              & $F^\omega_\beta \uparrow$               & MAE $\downarrow$           & $S_\alpha \uparrow$              & $E_\phi \uparrow$              & $F^\omega_\beta \uparrow$               & MAE $\downarrow$           \\ \hline
SINet\cite{SINet}                   & 0.869          & 0.891          & 0.740          & 0.440          & 0.751          & 0.771          & 0.606          & 0.100          & 0.771          & 0.806          & 0.551          & 0.051          \\
RankNet\cite{RankNet}                 & 0.846          & 0.913          & 0.767          & 0.045          & 0.712          & 0.791          & 0.583          & 0.104          & 0.767          & 0.861          & 0.611          & 0.045          \\
JCOD \cite{JCOD}                   & 0.870          & 0.924          & -              & 0.039          & 0.792          & 0.839          & -              & 0.82           & 0.800          & 0.872          & -              & 0.041          \\
PFNet \cite{PFNet}                  & 0.882          & 0.942 & 0.810          & 0.330          & 0.782          & 0.852          & 0.695          & 0.085          & 0.800          & 0.868          & 0.660          & 0.040          \\
FBNet  \cite{FBNet}                 & 0.888          & 0.939          & 0.828 & 0.032 & 0.783          & 0.839          & 0.702          & 0.081          & 0.809          & 0.889          & 0.684          & 0.035          \\
MM-SAM  \cite{FBNet}                 & 0.923          & 0.946          & 0.853 & 0.027 & 0.863          & 0.901          & 0.782          & 0.059          & 0.896          & 0.907          & 0.808          & 0.023          \\
SENet  \cite{FBNet}                 & 0.888          & 0.932          & 0.847 & 0.039 & 0.918          & 0.957          & 0.878          & 0.019          & 0.865          & 0.925          & 0.780          & 0.024          \\
\hline
SAM   \cite{SAM}    & 0.727 &  0.734   &  0.639    &  0.081  & 0.684 & 0.687 & 0.606 & 0.132  &  0.783    &    0.798     & 0.701 & 0.050   \\
SAM2   \cite{ravi2024sam2segmentimages}    & 0.359 &  0.375   &  0.115    &  0.357  & 0.350 & 0.411 & 0.079 & 0.311  &  0.429    &    0.505     & 0.115 & 0.218   \\
SAM-Adapter \cite{chen2023samfailssegmentanything,chen2023sam}                   & 0.896 & 0.919          & 0.824          & 0.033          & 0.847 & 0.873          & 0.765          & 0.070          & 0.883 & 0.918 & 0.801 & 0.025 \\
SAM2-Adapter \cite{chen2024sam2}                   & 0.915 & 0.955          & 0.889          & 0.018          & 0.855 & 0.909          & 0.810          & 0.051          & 0.899 & 0.950 & 0.850 & 0.018 \\
\hline
\textbf{SAM3-Adapter (Ours)}                    & \textbf{0.944} & \textbf{0.972}          & \textbf{0.908}          & \textbf{0.016}          & \textbf{0.919} & \textbf{0.954}          & \textbf{0.875}          & \textbf{0.029}          & \textbf{0.927} & \textbf{0.965} & \textbf{0.882} & \textbf{0.015} \\ \hline
    \end{tabular}}
\caption{Quantitative Segmentation Result Comparison for Camouflaged Object Detection}
\label{Quantitative_Segmentation_Result}
\end{table*}

\begin{figure*}[h]
\centering
\includegraphics[width=0.6\linewidth]{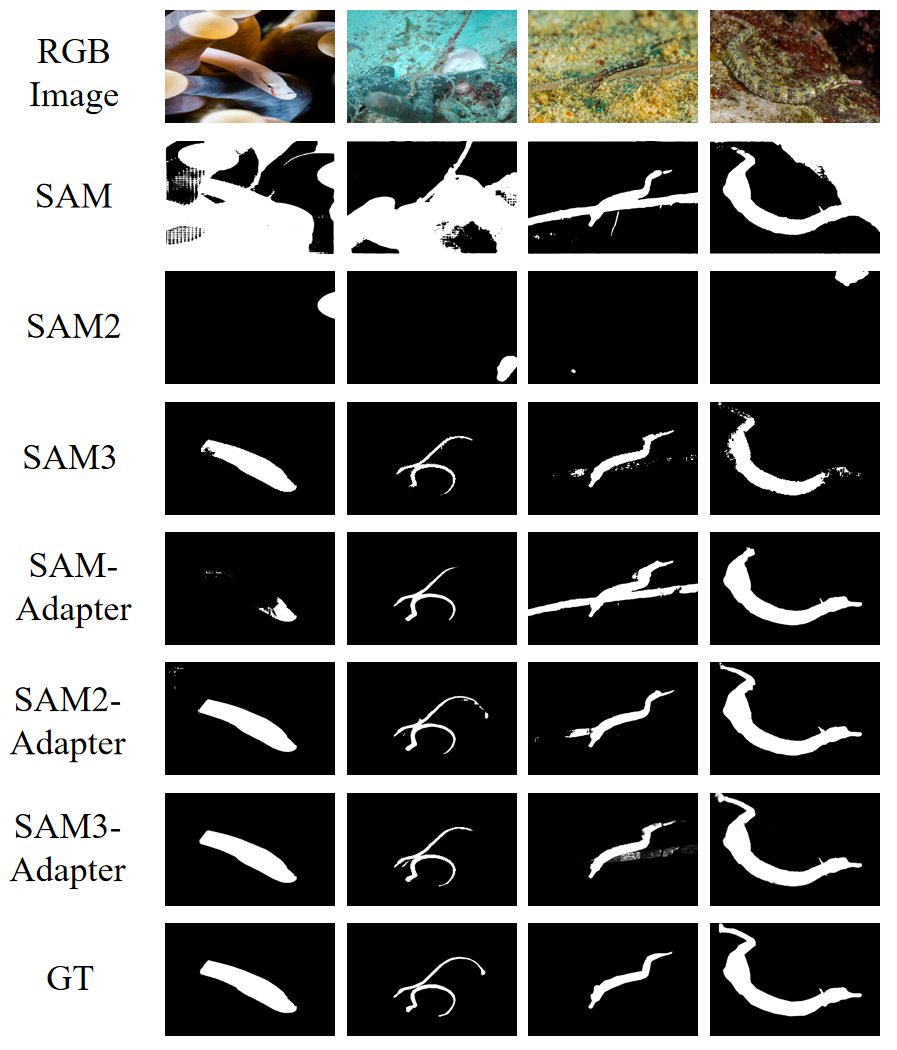}
\caption{\textbf{Camouflaged image segmentation on the COD-10K dataset.} Examples from the COD-10K dataset illustrating animals that are strongly camouflaged within their natural backgrounds. The original SAM frequently fails to accurately localize these targets and can produce fragmented or semantically incoherent segmentations; SAM2 and SAM3 exhibits similar limitations, occasionally producing no mask or incorrect outputs. With the integration of SAM3-Adapter, segmentation reliability on these challenging instances is substantially improved, achieving clear gains over earlier SAM2-Adapter variants.}
\label{fig3}
\end{figure*}

\begin{figure*}[h]
\centering
\includegraphics[width=0.75\linewidth]{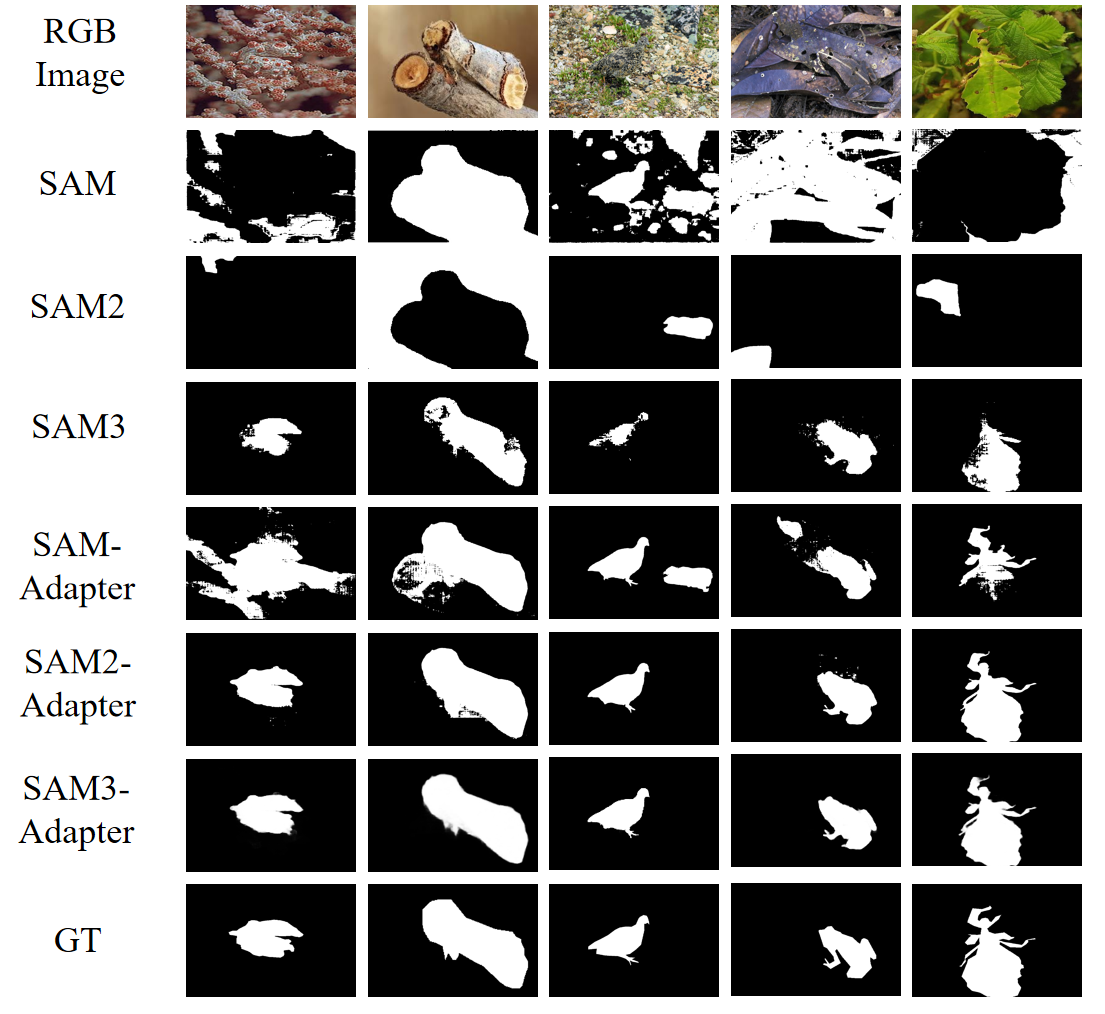}
\caption{\textbf{Camouflaged examples from the CAMO dataset.} The original SAM, SAM2 struggle to perceive animals that are visually concealed within their natural surroundings. SAM3, however, have already gained the capability of distinguish the camouflaged object as we can observe. Integrating SAM3-Adapter further improves the model's ability to segment the concealed targets.} \label{fig:4}
\end{figure*}

    The ${\rm MLP}_{tune}^i$ has one linear layer and ${\rm MLP}_{up}^i$ is one linear layer that maps the output from GELU activation to the number of inputs of the transformer layer. Balanced BCE loss is used for shadow detection. BCE loss and IOU loss are used for camouflaged object detection and polyp segmentation. The AdamW optimizer is used for all experiments. The initial learning rate is set to 2e-4. Cosine decay is applied to the learning rate. The batch size is 2. The training of camouflaged object segmentation is performed for 29 epochs. Shadow segmentation is trained for 29 epochs. Polyp segmentation is trained for 100 epochs. The experiments are implemented using PyTorch on NVIDIA Tesla A800 GPUs. For more information, please refer to the original SAM-Adapter paper \cite{chen2023samfailssegmentanything} and our codebase. Note that our codebase also supports Ascend 910b NPU.

\subsection{Experiments for Camouflaged Object Detection}

    We first evaluated the performance on the challenging task of camouflaged object detection, where objects are intentionally blended into their surroundings. Our initial analysis revealed that the powerful Segment Anything 3 (SAM3) model, on its own, already possesses a foundational capability to discern these concealed objects. Unlike its predecessors (SAM and SAM2) which often failed to produce meaningful results, SAM3 can typically locate the camouflaged targets. However, as shown in the baseline results in Figure \ref{fig2}, its segmentation masks often lack precision, with blurry boundaries and incomplete coverage. This is quantitatively reflected in Table \ref{Quantitative_Segmentation_Result}, where SAM3's standalone performance, while promising, does not yet match the state-of-the-art.
    
This is where SAM3-Adapter demonstrates its transformative impact. As vividly illustrated in Figures \ref{fig2}, \ref{fig3}, and \ref{fig:4}, the introduction of our adapter dramatically enhances SAM3's native ability. The segmentation results are not just improved; they are refined to a new level of accuracy. Our method produces masks with sharper, more precise contours that adhere tightly to the true object boundaries, effectively separating the camouflaged object from its visually similar background.

The quantitative results confirm this visual evidence. With the enhancement from SAM3-Adapter, our method not only surpasses the standalone SAM3 but also establishes a new state-of-the-art (SOTA) across all evaluated metrics. This significant performance leap, achieved by refining an already strong baseline, underscores the effectiveness of our adapter in unlocking and focusing the full potential of the SAM3 backbone for high-fidelity segmentation.

\subsection{Experiments for Shadow Detection}
\begin{figure*}[h]
\centering
\includegraphics[width=0.58\linewidth]{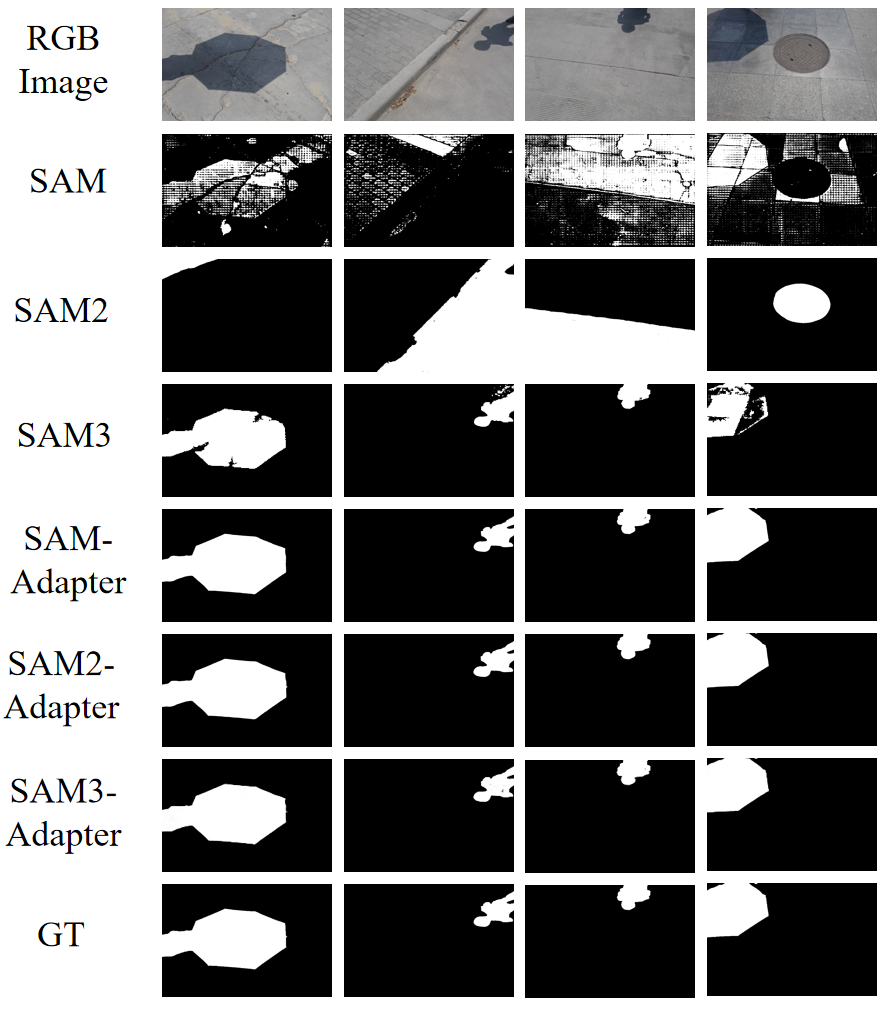}
\caption{\textbf{Visualization of Shadow Detection results.} SAM and SAM2 fails to identify shadows. Standalone SAM3 demonstrates a foundational ability to identify shadows, but struggles with precise boundaries. Our SAM3-Adapter unlocks SAM3's full potential, transforming its initial perception into state-of-the-art segmentation masks with sharp, accurate contours.}\label{fig5}
\end{figure*}

We extended our evaluation to the task of shadow detection, a challenge that requires discerning subtle, low-contrast regions from the background. Our analysis shows that the Segment Anything 3 (SAM3) model, on its own, demonstrates a clear, foundational understanding of the "shadow" concept. Unlike previous models that often failed entirely, SAM3 is generally able to identify the presence and approximate location of shadows in an image. However, as illustrated by the baseline results in Figure \ref{fig5}, these initial predictions often suffer from inaccuracies, such as incomplete segmentation, bleeding into non-shadow areas, and poorly defined edges.
This provides the perfect opportunity for SAM3-Adapter to showcase its value. By integrating our lightweight adapter, we transform SAM3's foundational capability into expert-level performance. The visual results in Figure \ref{fig5} are striking: where the standalone SAM3 produced ambiguous or noisy masks, our SAM3-Adapter method yields clean, precise shadow segmentations with sharp, well-defined contours. The adapter effectively teaches the model to respect the subtle boundaries of the shadow, eliminating the previously observed missing parts and erroneous additions.
The quantitative data presented in Table \ref{ablation_components} rigorously supports these visual improvements. The integration of SAM3-Adapter brings a significant performance boost, elevating the results far beyond the SAM3 baseline and setting a new state-of-the-art (SOTA) for shadow detection. This success further validates our core hypothesis: the most effective path to advancing segmentation is not just using a powerful backbone like SAM3, but synergistically enhancing it with intelligent, task-specific adapters.

\begin{table}[t]{
 \centering %
    \begin{tabular}{l|c c c}
    \toprule
    Method & BER $\downarrow$ \\
    \midrule
    Stacked CNN \cite{vicente2016large} & 8.60\\
    \midrule
    BDRAR \cite{BDRAR} & 2.69 \\
    \midrule
    DSC \cite{DSC} & 3.42\\
    \midrule
    DSD \cite{DSD} & 2.17 \\
    \midrule
    FDRNet \cite{zhu2021mitigating} & 1.55 \\
    \midrule
    SAM \cite{SAM} & 40.51\\
    SAM2 \cite{ravi2024sam2segmentimages} & 50.81\\
    SAM-Adapter  & 1.43\\
    SAM2-Adapter  & 1.43\\
    SAM3-Adapter (Ours) & \textbf{1.14}\\
     \bottomrule
    \end{tabular}
    \caption{Result for Shadow Detection}
    \label{ablation_components}}
\end{table}

\subsection{Experiments for Polyp Segmentation}
\begin{figure*}[h]
\centering
\includegraphics[width=0.57\linewidth]{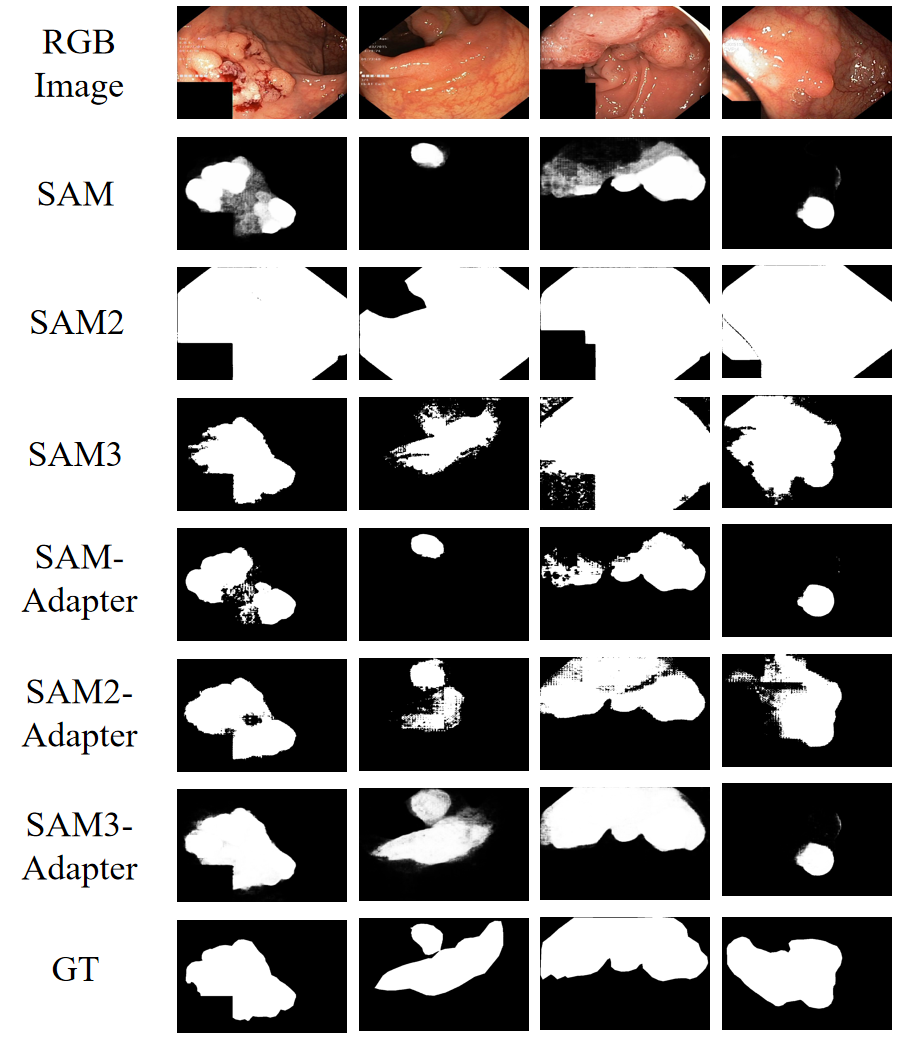}
\caption{\textbf{Qualitative results for Polyp Segmentation.} The figure illustrates that SAM struggles to accurately delineate polyp boundaries, and SAM2 produce non-meaningful outputs. While the powerful SAM3 model can successfully locate polyp tissues, its standalone segmentation often results in incomplete masks with poorly defined boundaries. Our SAM3-Adapter dramatically enhances this foundational capability, guiding the model to produce highly accurate and holistic segmentations. The resulting masks precisely delineate the entire polyp structure, significantly outperforming all baseline models \cite{chen2024xlstm,ma2024u} } \label{fig:6}
\end{figure*}

We then evaluated our method in the critical domain of medical image segmentation, specifically focusing on polyp segmentation. Accurate and reliable segmentation of polyps is paramount for the early detection and prevention of colorectal cancer, a leading cause of cancer-related deaths globally.

Our analysis begins by assessing the standalone capabilities of the Segment Anything 3 (SAM3) model. We found that SAM3, owing to its powerful architecture and vast pre-training, possesses a strong foundational ability to identify and locate polyps within colonoscopy images. This marks a significant improvement over prior generalist models. However, for a task where clinical precision is essential, SAM3's baseline performance reveals limitations: the resulting segmentation masks, while correctly positioned, often suffer from incomplete coverage and fuzzy boundaries, failing to capture the entire polyp structure accurately (as shown in the baseline examples in Figure \ref{fig:6}).

This is precisely the gap that SAM3-Adapter is designed to fill. By integrating our efficient adapter, we elevate SAM3's performance from foundational to state-of-the-art. The visual results, presented in Figure \ref{fig:6}, are compelling. Our method transforms SAM3's initial, coarse predictions into highly accurate, holistic segmentation masks that precisely delineate the full contour of the polyp tissue. The adapter effectively channels SAM3's powerful features to focus on the subtle yet critical details required for medical-grade accuracy.

The quantitative results in Table \ref{fd} provide definitive proof of this superiority. The combination of SAM3 and SAM3-Adapter not only dramatically outperforms the standalone SAM3 but also establishes a new state-of-the-art (SOTA), surpassing all previous methods. This achievement underscores the immense value of our approach: by unlocking and refining the power of a scaled-up foundation model, we provide a more accurate and reliable tool for a crucial clinical application.

\begin{table*}[t]{
     \centering %
    \begin{tabular}{l | c c c}
    \toprule
    Method & mDice $\uparrow$ & mIoU $\uparrow$\\
    \midrule
    UNet \cite{unet} & 0.821 & 0.756\\
    \midrule
    UNet++ \cite{zhou2018unet++} & 0.824 & 0.753 \\
    \midrule
    SFA \cite{fang2019selective} & 0.725 & 0.619 \\
    \midrule
    SAM \cite{SAM} & 0.778 & 0.707\\
    SAM2 \cite{ravi2024sam2segmentimages} & 0.200 & 0.029\\
    SAM-Adapter  & 0.850 & 0.776\\
    SAM2-Adapter  & 0.873 & 0.806\\
    SAM3-Adapter (Ours) & \textbf{0.906} & \textbf{0.842}\\
     \bottomrule
    \end{tabular}

    \caption{Quantitative Result for Polyp Segmentation}
    \label{fd}}
\end{table*}

\subsection{Experiments for Cell Segmentation}
To further test the generalizability of our approach, we applied it to the highly challenging task of cell segmentation. This domain demands extreme precision to distinguish individual cells in dense, often overlapping clusters. As evidenced by both qualitative results and quantitative metrics in Table \ref{table_cell_seg}, our method achieved a staggering improvement over all previous state-of-the-art methods. The performance leap was more substantial here than in any other downstream task that we evaluated, showcasing the immense potential of our approach for biomedical research and diagnostics.

\begin{table}[h!]
\centering
\caption{F1 Score Comparison on Cells in Microscopy}
\begin{tabular}{l c}
\hline
Methods & F1 $\uparrow$ \\
\hline
nnU-Net        & 0.5383 \\
SegResNet      & 0.5411 \\
UNETR          & 0.4357 \\
SwinUNETR      & 0.3967 \\
U-Mamba\_Bot   & 0.5389 \\
U-Mamba\_Enc   & 0.5607 \\
xLSTM-UNet\_bot      & 0.5818 \\
xLSTM-UNet \_enc      & 0.6036 \\
\hline
\textbf{SAM3-Adapter (Ours)} & \textbf{0.7525} \\
\hline
\end{tabular}
\label{table:f1_only}
\end{table}

\section{Conclusion and Future Work}
    In this work, we presented SAM3-Adapter, a parameter-efficient tuning method that elevates Segment Anything 3 (SAM3) to a new level of performance for specialized segmentation. Through comprehensive experiments, we have demonstrated that SAM3-Adapter sets a new state-of-the-art (SOTA) in difficult segmentation tasks such as medical, camouflage, and shadow detection. Crucially, it achieves these results with higher computational efficiency. 

    Our extensive experiments validate this hypothesis. By synergizing our lightweight adapter with the powerful SAM3 backbone, we have established new state-of-the-art (SOTA) benchmarks in challenging domains like medical, camouflage, and shadow segmentation. This combination not only surpasses previous SAM2-based methods in accuracy but also does so with remarkable parameter efficiency, showcasing the tangible benefits of SAM3's advanced design when properly adapted.
    
The success of SAM3-Adapter provides a clear demonstration: scaling up foundation models, when paired with intelligent and efficient adaptation techniques, directly translates to significant gains in specialized, real-world applications. Our method acts as a catalyst, enabling the powerful general representations learned by SAM3 to be effectively channeled into high-fidelity, domain-specific segmentation.

We advocate for the adoption of SAM3, amplified by our SAM3-Adapter, as the new frontier for high-performance image segmentation. We are releasing our code, models, and protocols to empower the community to build upon this powerful synergy and drive further progress in adapting large-scale vision models. Code, pre-trained models, and data processing protocols are available at \href{http://tianrun-chen.github.io/SAM-Adaptor/}{http://tianrun-chen.github.io/SAM-Adaptor/}

{
    \small
    \bibliographystyle{ieeenat_fullname}
    \bibliography{main}
}

\end{document}